\documentclass[letterpaper, 10 pt, conference]{ieeeconf}  

\IEEEoverridecommandlockouts                              
\usepackage{amsmath}
\usepackage{amssymb}
\usepackage{amsfonts}
\usepackage[pdftex]{graphicx}
\DeclareGraphicsExtensions{.pdf, .png, .jpg}
\usepackage{algorithm}
\usepackage[noend]{algpseudocode}
\algrenewcommand\algorithmicindent{0.7em}%
\usepackage{gensymb}
\usepackage[caption=false]{subfig}
\usepackage{cite}
\usepackage{hyperref}
\usepackage{flushend}

\usepackage{amsthm}

\newtheorem{theorem}{Theorem}
\theoremstyle{definition}

\newtheorem{lemma}{Lemma}

\DeclareMathOperator*{\E}{\mathbb{E}}
\DeclareMathOperator*{\argmax}{\text{argmax}}

\usepackage[usenames,dvipsnames]{xcolor}

\title{\LARGE \bf
An Upper Confidence Bound for Simultaneous Exploration and Exploitation in Heterogeneous Multi-Robot Systems
}

\author{Ki Myung Brian Lee$^1$, Felix Kong$^1$, Ricardo Cannizzaro$^2$, Jennifer L. Palmer$^2$, \\
David Johnson$^3$, Chanyeol Yoo$^1$ and Robert Fitch$^1$ 
\thanks{This work is supported by an Australian Government Research Training Program (RTP) Scholarship, Australia's Defence Science and Technology Group, and the University of Technology Sydney.}
\thanks{$^1$Authors are with the University of Technology Sydney, Ultimo, NSW 2006, Australia {\tt\footnotesize brian.lee@student.uts.edu.au, \{chanyeol.yoo, felix.kong, rfitch\}@uts.edu.au}}
\thanks{$^2$Authors are with the Defence Science and Technology Group, Department of Defence, Australia {\tt\footnotesize \{ricardo.cannizzaro, jennifer.palmer\}@dst.defence.gov.au}}
\thanks{$^3$Author is with Mission Systems Pty. Ltd., Sydney, Australia {\tt\footnotesize david.johnson@missionsystems.com.au}}
}

\begin{document}

\maketitle

\begin{abstract}
Heterogeneous multi-robot systems are advantageous for operations in unknown environments because functionally specialised robots can gather environmental information, while others perform tasks. We define this decomposition as the \emph{scout\textendash task robot architecture} and show how it avoids the need to explicitly balance exploration and exploitation~by permitting the system to do both simultaneously. The challenge is to guide exploration in a way that improves overall performance for time-limited tasks. We derive a novel upper confidence bound for simultaneous exploration and exploitation based on mutual information and present a general solution for scout\textendash task coordination using decentralised Monte Carlo tree search. We evaluate the performance of our algorithms in a multi-drone surveillance scenario in which scout robots are equipped with low-resolution, long-range sensors and task robots capture detailed information using short-range sensors. The results address a new class of coordination problem for heterogeneous teams that has many practical applications.
\end{abstract}

\section{Introduction}
Multi-robot systems enable flexible scaling of robotic applications by composing multiple, possibly disposable robots into a functional team that outperforms a single robot. 
Real-world use of multi-robot systems is increasing, for example, in warehouse management \cite{amazon}, agriculture \cite{swarmfarm}, and defence \cite{a2ad-with-swarming}. 
We anticipate that heterogeneity will further accelerate the adoption of multi-robot systems because it enables increases in system capability through functional specialisation of individual robots. Specialisation is particularly appealing in applications in which the environment is partially or completely unknown. A subset of the team could focus on gathering information, while the rest perform tasks. We are interested in developing algorithms to coordinate the behaviour of heterogeneous teams that operate in unknown environments and in exploring how the division of labour between information-gathering and task-performing robots relates to the classical trade-off between exploration and exploitation.

\begin{figure}[t!]
    \centering
    \includegraphics[width=0.97\columnwidth]{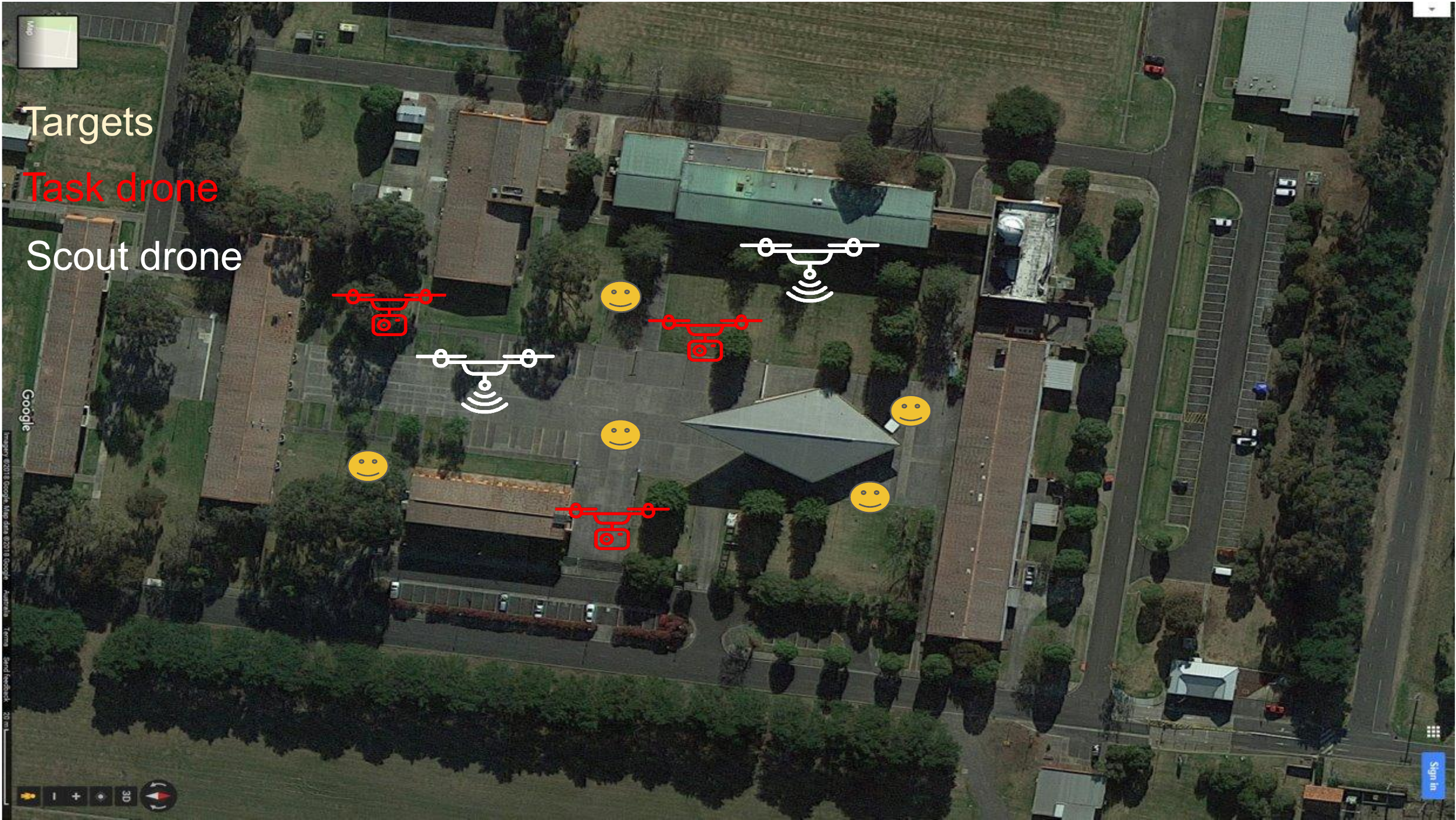}
    \caption{An example application for multi-drone surveillance. The task is to confirm all targets (yellow) with the task drones (red). Scout drones (white) support the process by sensing targets (yellow) from a distance, at low resolution, and cueing for possible target presence.\label{fig:scenario}}
\vspace{-2ex}
\end{figure}

We define a team composition in which some robots (i.e., \emph{task robots}) are equipped to perform a particular task while others (i.e., \emph{scout robots}) are equipped with sensors to rapidly acquire knowledge about the environment as the \emph{scout\textendash task robot architecture}.
There are many compelling applications of this idea.
For example, it may be desirable to deploy disposable scout robots to ensure safe operation of a high-value task robot, as in the case of a Mars rover-copter team \cite{ali_agha_mohamadi_where_to_map,mohammadi_2,takashi_tanaka}.
We consider the multi-drone surveillance application illustrated in Fig.~\ref{fig:scenario}, where a limited number of scout robots equipped with long-range sensors cue task robots for the presence of targets. 

One unexplored benefit of the scout\textendash task robot architecture is that it can be viewed as a way to sidestep the exploration\textendash exploitation trade-off inherent to operations in unknown environments, through heterogeneity. Whereas the trade-off between exploration and exploitation is a question of balancing these two activities, the scout\textendash task architecture permits simultaneous exploration and exploitation. Assuming that the overall system objective is to complete as many tasks as possible in a given amount of time, the challenge is how to guide exploration in a way that is most relevant to exploitation. Scout robots should provide information about the environment that allows task robots to improve their plans and thus find higher-quality solutions.
For this reason, we find that the scout\textendash task coordination problem is fundamentally different from previous heterogeneous multi-robot coordination problems, and solutions have been proposed only in application-specific instances~\cite{ali_agha_mohamadi_where_to_map,takashi_tanaka,mohammadi_2}.

In this paper, we present a general solution to scout\textendash task coordination. We derive a novel upper confidence bound (UCB), the \textit{mutual-information UCB} (MI-UCB), to enable simultaneous exploration and exploitation. 
The MI-UCB shows that the posterior expected reward is probabilistically upper bounded by a combination of Shannon information gain and prior expected reward.
It then follows from the principle of optimism under uncertainty that executing paths that maximise the MI-UCB, in fact, maximises the posterior expected reward in hindsight. 
We apply MI-UCB to the multi-drone surveillance problem shown in Fig.~\ref{fig:scenario} using decentralised Monte Carlo tree search (Dec-MCTS) \cite{decmcts}. 
Our results show that, with the same team configuration, the hindsight reward is improved by up to 134\% compared with simply maximising prior expected reward.
We also demonstrate MI-UCB in a multi-drone simulation with real-time operations set in a realistic environment. 

\section{Related Work}
Heterogeneous multi-robot coordination is often posed as a task-allocation problem in which robots have varying levels of competency in completing each task.
An optimal assignment of tasks may be achieved with, e.g., markets \cite{markets}, Hungarian algorithms \cite{hungarian}, or MCTS \cite{smith_and_hollinger}. 
We are interested in the scout\textendash task robot coordination problem, where a sub-team of scout robots assists task robots in completing their tasks by exploring the environment. 
Exploration implicitly aids task completion in the long term. 
Existing work closely related to this problem includes\textcolor{red}{} \cite{ali_agha_mohamadi_where_to_map,takashi_tanaka}, which considers a Mars rover completing navigation or temporal-logic tasks, assisted by an aerial robot that scouts ahead to improve localisation accuracy or environmental knowledge. 
While we do not consider these problem instances in this paper, our result is sufficiently general to encompass them.  

Multi-robot surveillance, in which a team of robots searches for targets \cite{hollinger_survey}, is typically addressed by maintaining a probabilistic belief over target locations using an occupancy grid \cite{vidal_pursuit_evasion} or a random finite set \cite{clark2006gm,dames,Sung2018}. 
Based on the current belief, a plan is generated that maximises the expected number or probability of detections \cite{pursuit_evasion_uav_ugv,optimal_search_for,hollinger_multi_robot_moving,hollinger_survey,seng_keat_gan_collision} or minimises the uncertainty of target locations \cite{brent_schotfeldt,cliff2018robotic,dames}.
Here, the two objectives are distributed across the team, such that scout robots contribute to uncertainty reduction and task robots to detection of targets. 
We show that considering the two objectives in tandem improves the overall number of detections \emph{in hindsight}, compared with considering only the expected number of detections. 

The idea of combining exploration and exploitation is not new. 
In martingale-based approaches such as UCB on trees (UCT) \cite{auer} or KL-UCB \cite{kl_ucb}, an agent repeatedly samples the reward of an action to construct statistical quantities that motivate exploration or exploitation. 
A trade-off between the two gradually biases the sampling toward the optimum, as in the case of MCTS \cite{decmcts,davidsilver}. 
These approaches are parallel to ours, because it is infeasible for a physical robot to sufficiently sample reward during its operation.

A more suitable class of algorithms for robotic applications in unknown environments is Bayesian optimisation (BO). 
A prominent example is Gaussian process (GP) UCB \cite{gp_ucb}, which has been used in robotic source seeking in plumes \cite{brian_acra} and in human\textendash robot interaction \cite{sungjoon_choi}.
In BO, an agent cycles between gathering a new sample and updating its belief; and the samples are biased toward the optimum by use of a UCB derived from that belief.
This is prohibitive for scout\textendash task coordination, as all agents must necessarily contribute to both gathering measurements and maximising reward. 
Our proposed UCB substantially relaxes this requirement and others using information-theoretic tools. 

A theoretical result closely related to our work is \cite[Lemma 3]{stanford_guys}, which shows that, if a UCB similar to ours were to hold, using the UCB as an acquisition function for action selection leads to bounded regret. 
We show that a similar UCB generally holds.
\section{Scout-Task Coordination Problem}

\begin{figure}[tb]
    \centering
    \includegraphics[width=\columnwidth]{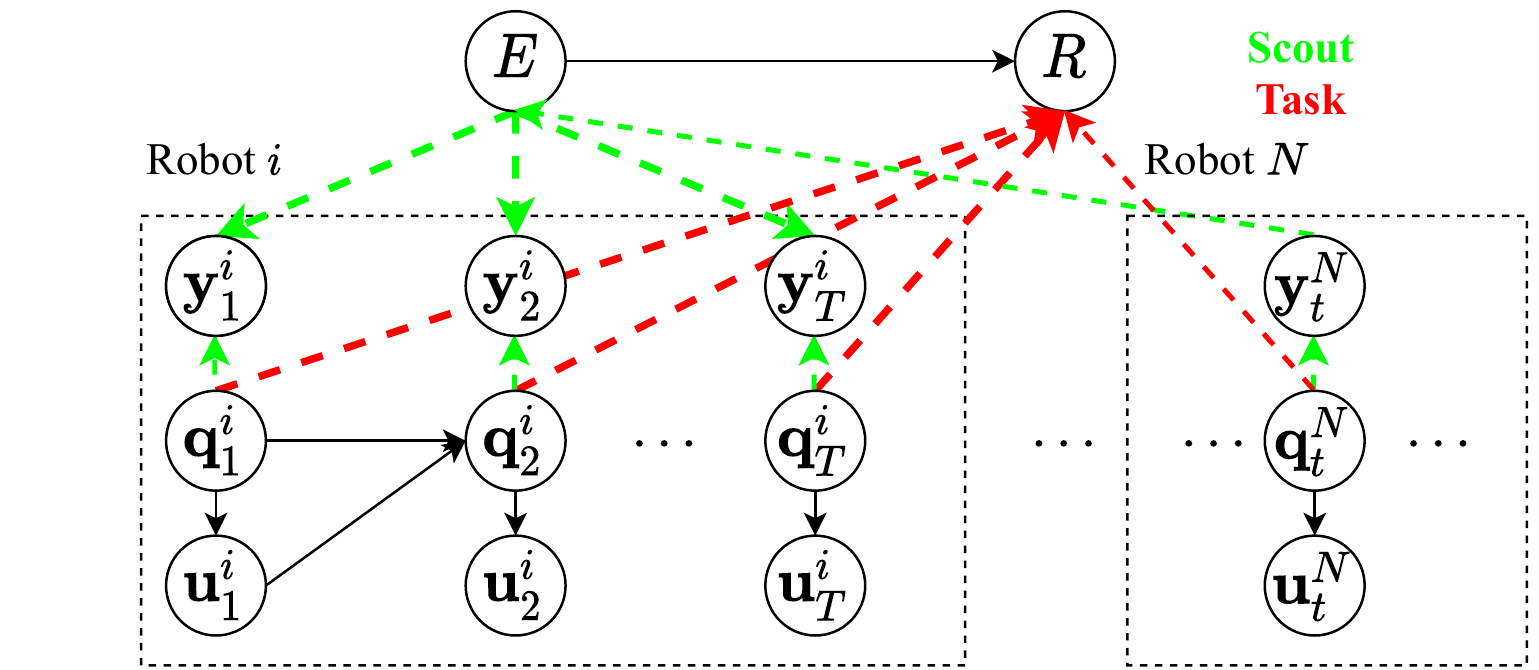}
    \caption{
    A probabilistic graphical model illustrating the scout\textendash task coordination problem. 
    Dashed connections depend on team composition.
    The control input $\mathbf{u}$ generates trajectory $\mathbf{q}$. 
    A task robot (red) gains a reward $R(\mathbf{q}, E)$, depending on the trajectory $\mathbf{q}$ and the latent environment $E$. 
    A scout robot (green) gathers measurements $\mathbf{y}_{t}$ at each state $\mathbf{q}_{t}$, revealing information about $E$. 
    \label{fig:pgm}}
\vspace{-1ex}
\end{figure}
We formulate the scout\textendash task coordination problem, illustrated in Fig.~\ref{fig:pgm}, as follows. 
Consider a team of $N$ mobile robots, the dynamics of which are described by:
\begin{equation}
\label{eq:dyn}
    \mathbf{q}_{t+1}^{r} = \mathbf{f}(\mathbf{q}_{t}^{r}, \mathbf{u}_{t}^{r}), 
\end{equation}
where $\mathbf{q}_{t}^{r}$ and $\mathbf{u}_{t}^{r}$ are the state and control action of robot $r$ at time $t$,  respectively. 
The superscript $1 \leq r \leq N$ denotes the robot, the subscript $t$ denotes time, and $\mathbf{u}_{t}^{r}$ is the control action applied to robot $r$ at time $t$. 

The robots operate in an unknown environment, denoted by $E$, which can follow any distribution. 
For example, it may be discrete and follow a categorical distribution; or it may be continuous and follow a Gaussian distribution. 
Each robot may belong to a set of scout robots $\mathcal{S} \subset [1, ..., N]$ or to a set of task robots $\mathcal{T} \subset [1, ..., N]$.
$\mathcal{S}$ and $\mathcal{T}$ are not necessarily disjoint; and thus a robot may belong to both sets (i.e., it may be a scout-and-task robot).

If $r \in \mathcal{S}$, robot $r$ generates measurements $\mathbf{y}_{t}^{r}$ that reveal information about $E$:
\begin{equation}
    \mathbf{y}_{t}^{r} \sim \mathcal{P}(\mathbf{y}_{t}^{r} \mid \mathbf{q}_{t}^{r}, E). 
\end{equation}
If $r \in \mathcal{T}$, robot $r$ is equipped with a payload to perform an intended task, which depends on the environment $E$. 
We thus model the task completion by a deterministic reward function $R(\mathbf{q}^{\mathcal{T}}, E)$. 
It is important to note that scout-only robots $r\in\mathcal{S} \setminus \mathcal{T}$ do not contribute directly to the reward function; instead they allow the task and scout-and-task robots $r\in\mathcal T$ to more effectively complete their tasks by gathering information on $E$.

For brevity, we omit the subscript (resp. the superscript) to mean the set of states over time (resp. different robots).
I.e., $\mathbf{q}^{r} = \{ \mathbf{q}_{1}^{r}, ..., \mathbf{q}_{T}^{r}\} $, $\mathbf{q}_{t} = \{ \mathbf{q}_{t}^{1}, ..., \mathbf{q}_{t}^{N} \}$.
The omission of both subscript and superscript indicates the set of all robots' trajectories over time: i.e., $\mathbf{q} = \{ \mathbf{q}^{1}, ... , \mathbf{q}^{N} \} = \{ \mathbf{q}_{1}, ..., \mathbf{q}_{T} \}$. 
We also replace the superscript with $\mathcal{S}$ and $\mathcal{T}$ to mean the set of poses or trajectories of robots that belong to the set of scout or task robots, respectively. 
Hence, $\mathbf{q}^{\mathcal{T}} = \{ \mathbf{q}^{i} \mid i \in \mathcal{T} \}$.
Further, we write $\mathbf{q^{r}}(\mathbf{u}^{r})$ to mean the trajectory obtained by applying $\mathbf{u}^{r}$ to robot $r$; and likewise $\mathbf{q}(\mathbf{u})$ denotes the set of all robots' trajectories obtained by applying the control sequences $\mathbf u = \{\mathbf u^1,\ldots,\mathbf u^r\}$ to the corresponding robots. 
Similarly, we write $\mathbf{y}^{r}(\mathbf{q}^{r})$ to mean the set of observations obtained from the trajectory of robot $r$. 

Our aim is to choose control inputs $\mathbf{u}$ maximising reward:
\begin{equation}\label{eq:argmax}
    \mathbf{u^{*}} = \argmax_{\mathbf{u}^\mathcal{T}} R(E, \mathbf{q}^{\mathcal{T}}(\mathbf{u}^{\mathcal{T}})).
\end{equation}
Clearly, \eqref{eq:argmax} cannot be solved directly because $E$ is a random variable and, consequently, so is $R(E, \mathbf{q}^{\mathcal{T}}(\mathbf{u}^{\mathcal{T}}))$. 
This is addressed in Sec.~\ref{sec:mi-ucb}.

\section{MI-UCB for Coordination}\label{sec:mi-ucb}

In this section, we derive a surrogate acquisition function, MI-UCB, by analysing the effect of improvement in environmental knowledge on the estimated reward. 
Then, we present an online planning framework for scout\textendash task coordination. 

\subsection{Mutual-Information Upper Confidence Bound (MI-UCB)} 

We cannot solve \eqref{eq:argmax} directly as $E$ and $R(\mathbf{q}(\mathbf{u}), E)$ are random variables that must be estimated. 
A common approach is to pose an \emph{expectimax} problem based on current belief:
\begin{equation}\label{eq:expectimax}
    \mathbf{u}^{*} = \argmax_{\mathbf{u}} \E_{E \sim \mathcal{P}(E)}  R(E, \mathbf{q}^{\mathcal{T}}(\mathbf{u}^{\mathcal{T}})).
\end{equation}
However, this does not show how, or why, the scout robots can coordinate because neither the measurements $\mathbf{y}^{\mathcal{S}}$, nor the state or control, $\mathbf{q}^{\mathcal{S}}$ or $\mathbf{u}^{\mathcal{S}}$, respectively, appear in \eqref{eq:expectimax}.

To consider the effect of measurements obtained by the scout robots, we consider maximising the posterior expected reward given measurements:
\begin{equation}\label{eq:posterior_expectimax}
    \mathbf{u}^{*} = \argmax_{\mathbf{u}} \E_{E \sim \mathcal{P}(E \mid \mathbf{y}^{\mathcal{S}})} R(E, \mathbf{q}^{\mathcal{T}}(\mathbf{u}^{\mathcal{T}})) .
\end{equation}
While \eqref{eq:posterior_expectimax} incorporates the effect of measurements, we cannot solve it directly because the measurements $\mathbf{y}^{\mathcal{S}}$ remain random variables that have not been sampled. 
One may take an expectation over possible measurements similar to a partially observable Markov decision process approach \cite{despot}, but the fundamental challenge of enumerating possible measurements over possible trajectories remains.

Our finding is that we can solve \eqref{eq:posterior_expectimax} using the \emph{principle of optimism under uncertainty}, by deriving a UCB on the posterior expected reward. 
The exact statement is:
\begin{theorem}[MI-UCB]\label{thm:post_prior}
Suppose $R(E, \mathbf{q}^{\mathcal{T}})$ is a measurable function of $E$ for all $\mathbf{q}^{\mathcal{T}}$. With probability $\geq 1 - \delta$: 
\begin{equation}\label{eq:post_prior}\begin{aligned}
    \E_{E \sim \mathcal{P}(E \mid \mathbf{y}^{S})}&[R(E, \mathbf{q}^{\mathcal{T}})] \leq \\ &\frac{1}{\delta} I(\mathbf{y}^{S}; E)  + \log \mathbb{E}[\exp R(E, \mathbf{q}^{\mathcal{T}}))]. 
\end{aligned}
\end{equation}
\end{theorem}
Most importantly, the UCB on the RHS of \eqref{eq:post_prior} \emph{decouples} scout- and task-only robots because it separates information gain and reward into a weighted sum.
The term $\log \E [\exp R(E, \mathbf{q}^{\mathcal{T}}(\mathbf{u}^{\mathcal{T}}))]$ is called the cumulant generating function (CGF) in probability theory. for which analytical expressions are often available. 
It is also important to note that the posterior expected reward on the left-hand side of \eqref{eq:post_prior} cannot be calculated before taking the measurement, 
while the UCB on the right-hand side may be calculated beforehand.

Theorem \ref{thm:post_prior} is proved by evaluating how the change from $\mathcal{P}(E)$ to $\mathcal{P}(E \mid \mathbf{y}^{S})$ affects the estimated reward, as captured by the seminal result of Donsker \& Varadarhan \cite{donsker_and_varadhan,dupuis,seldin}: 
\begin{lemma}[Change of measure inequality]\label{lem:donsker_varadhan}
Given any measurable function $\phi$ on $X$ and any two distributions $\mathcal{P}$ and $\mathcal{Q}$ on $X$, we have:
\begin{equation}
    \E_{x \sim P}[ \phi(x) ] \leq D_{KL}(\mathcal{P} \mid \mathcal{Q}) + \log \E_{x \sim Q}[ \exp\phi(x)]. 
\end{equation}
\end{lemma}

\begin{proof}[Proof of Theorem \ref{thm:post_prior}]
Consider the change of measure inequality between $\mathcal{P}(E \mid \mathbf{y}^{S})$ and $\mathcal{P}(E)$:
\begin{equation}
\begin{aligned}
    \E_{E \sim \mathcal{P}(E \mid \mathbf{y}^{S})}[R(E, \mathbf{q}^{\mathcal{T}})] \leq &D_{KL}(\mathcal{P}(E \mid \mathbf{y}^{S}) \mid \mathcal{P}(E)) \\
    + &\log \E [\exp R(E, \mathbf{q}^{\mathcal{T}})].
\end{aligned}
\end{equation}
Applying Markov's inequality over $\mathbf{y}^{\mathcal{S}}$ to the KL divergence term yields the claimed result.
\end{proof}

\subsection{Online Planning}

Based on the UCB in Theorem~\ref{thm:post_prior}, our online planning framework solves the following surrogate problem:
\begin{equation}\label{eq:argmax_ucb}
    \mathbf{u}^{*} = \argmax_{\mathbf{u}} I(E; \mathbf{y}^{\mathcal{S}} ) + \delta \log \E \exp R(E, \mathbf{q}^{\mathcal{T}}(\mathbf{u}^{\mathcal{T}})). 
\end{equation}
The principle of optimism under uncertainty \cite{brafman} asserts that maximising the UCB \eqref{eq:argmax_ucb} maximises the reward function \eqref{eq:posterior_expectimax} when evaluated in hindsight. 
This is known as a \emph{no-regret} bound. 


The online-planning framework cycles between updating the belief $\mathcal{P}(E)$ and maximising the MI-UCB \eqref{eq:argmax_ucb}.
For this purpose, we use Dec-MCTS~\cite{decmcts}, a decentralised multi-robot planning algorithm that extends the well-known MCTS.
As we do not modify Dec-MCTS except the objective function, we only give a brief description -- interested readers are referred to~\cite{decmcts}.

In Dec-MCTS, each robot maintains a probability distribution over the control sequences $\mathbf{u}$ of the entire team. 
Other robots' distributions are updated asynchronously via communication, while each robot's own distribution is updated via single-robot MCTS iterations. 
The single-robot MCTS iterations generate rollout trajectories for a single robot, and evaluates the objective function~\eqref{eq:argmax_ucb} with other robots' trajectories fixed at a random sample drawn from the probability distribution maintained.

Hence, the combination of MI-UCB and Dec-MCTS can solve any instance of the scout-task coordination problem in a decentralised manner as long as the MI-UCB~\eqref{eq:argmax_ucb} can be computed given the control sequences $\mathbf{u}$ of the entire team.
As distribution updates are asynchronous by design, Dec-MCTS is robust against delays, and thus permits multi-hop communication.

\section{Application in Multi-Drone Surveillance}
We apply MI-UCB to a multi-drone surveillance problem in which the task is to maximise the number of confirmations of an unknown number of targets at unknown locations. 
Task drones are equipped with short-range sensors only, and scout drones are equipped with long-range sensors that can rapidly provide knowledge about the environment. Drones may also be dual-equipped (i.e., they may be scout-and-task drones).

\subsection{Reward Function}
We represent the targets in a 2D occupancy grid, so that the environment is a Boolean matrix $E \in \mathbb{B}^{N_{X} \times N_{Y}}$, where $N_{X}$ and $N_{Y}$ are the number of cells in the $X$ and $Y$ directions, respectively. 
$E(i,j) = 1$ means cell $(i,j)$ is occupied by a target, and $0$ indicates otherwise. 

We model the visibility of cell $(i,j)$ from robot $r$ at time $t$ as a Bernoulli random variable $v_{t}^{r}(i,j; \mathbf{q}_{t}^{r}) \in \mathbb{B}$:
\begin{equation}
    v_{t}^{r}(i,j; \mathbf{q}_{t}^{r}) \sim \mathcal{P}(v_{t}^{r}(i,j) \mid \mathbf{q}_{t}^{r}).
\end{equation}
The visibility over a trajectory $\mathbf{q}^{r}$ is a disjunction $v^{r}(i,j; \mathbf{q}^{r}) = \vee_{t} v_{t}^{r}(i,j; \mathbf{q}^{r}_{t})$. Similarly for the visibility over different robots, $v(i,j; \mathbf{q}) = \vee_{r} v^{r}(i,j; \mathbf{q}^{r})$. Robot $r \in \mathcal{T}$ confirms a target at cell $(i,j)$ iff the target exists and may be sensed. The reward is the number of targets confirmed:
\begin{equation}\label{eq:surveillance_reward}
    R(\mathbf{q}^{\mathcal{T}}(\mathbf{u}^{\mathcal{T}}), E) = \sum_{ij} (v_{t}^{r}(i,j; \mathbf{q}^{\mathcal{T}}(\mathbf{u}^{\mathcal{T}}))E(i,j)). 
\end{equation}

The reward function \eqref{eq:surveillance_reward} is a sum of Bernoulli random variables, which is in turn a Poisson binomial random variable. 
Its CGF is given by:
\begin{equation}
\begin{aligned}
\log &\E_{E \sim \mathcal{P}(E)} \exp R(\mathbf{q}^{\mathcal{T}}(\mathbf{u}^{\mathcal{T}}), E) \\
&= \sum_{ij} \log(1 + \mathcal{P}(d(i,j; \mathbf{q}^{\mathcal{T}}(\mathbf{u}^{\mathcal{T}}))) (e - 1) ),
\end{aligned}
\end{equation}
where $\mathcal{P}(d(i,j; \mathbf{q}^{\mathcal{T}}(\mathbf{u}^{\mathcal{T}}))) = \mathcal{P}(v(i,j; \mathbf{q}^{\mathcal{T}}(\mathbf{u}^{\mathcal{T}}))) \mathcal{P}(E(i,j))$.

\subsection{Belief Update and Information Gain}
We use a simple grid-based filter for decentralised data fusion of $E$. 
With the standard independence assumption, the belief over target occupancy decomposes as:
\begin{equation}
\mathcal{P}(E) = \prod_{i,j} \mathcal{P}(E(i,j)).
\end{equation}

When a target is visible, a scout robot can measure its position. 
We adopt the inverse sensor model \cite{sebastian_thrun} approach to discretise the measurements and represent the measurement as a matrix of Bernoulli random variables:  
\begin{equation}\begin{aligned}
    \mathcal{P}(\mathbf{y}^{r}_{t}(i,j) &\mid E(i,j), \mathbf{q}_{t}^{r}) = \\ &v_{t}^{r}(i,j ; \mathbf{q}_{t}^{r}) \mathcal{P}(\mathbf{y}^{r}_{t}(i,j) \mid E(i,j) ).
\end{aligned}\end{equation}
The sensor model $\mathcal{P}(\mathbf{y}^{r}_{t}(i,j) \mid E(i,j))$ is given by a confusion matrix between true and measured occupancy. 

Each scout robot communicates its position and detected target locations (if any) at regular intervals. 
Measurements are fused with Bayes' rule:
\begin{equation}
\begin{aligned}
    \mathcal{P}(E(i,j&)\mid \mathbf{y}^{r}_{1:t}(i,j) ) =  \Big((1 - v_{t}^{r}(i,j ; q_{t}^{r})) \\  
              + & v_{t}^{r}(i,j ; q_{i}^{r}) \frac{\mathcal{P}(\mathbf{y}^{r}_{t}(i,j) \mid E(i,j), \mathbf{y}^{r}_{1:t-1}(i,j)  }{\mathcal{P}(\mathbf{y}^{r}_{t}(i,j) \mid \mathbf{y}^{r}_{1:t-1}(i,j) }\Big) \\
\times & \mathcal{P}(E(i,j)\mid \mathbf{y}^{r}_{1:t-1}(i,j) ).
\end{aligned}
\end{equation}

Information gain may be calculated as follows.  
For each cell, the information gain is:
\begin{equation}\begin{aligned}
    I(E&(i,j); \mathbf{y}^{r}_{t}(i,j))) =  \\
    &H( \mathcal{P}(\mathbf{y}^{r}_{t}(i,j)))  -  \E_{E(i,j)}H(\mathcal{P}(\mathbf{y}^{r}_{t}(i,j) \mid E(i,j))),
\end{aligned}\end{equation}
where $H(p)$ is binary entropy and $\mathcal{P}(\mathbf{y}^{r}_{t}(i,j)) = \E_{E(i,j)} \mathcal{P}(\mathbf{y}^{r}_{t}(i,j) \mid E(i,j))$.
The information gain is summed over the visible region: 
\begin{equation}
\begin{aligned}
    I(E; \mathbf{y}^{r}_{t}) = \sum_{ij} v(i,j; \mathbf{q}(\mathbf{u})) I(E(i,j); \mathbf{y}^{r}_{t}(i,j))).
\end{aligned}
\end{equation}
\section{Results}
We analyse the performance of MI-UCB in the context of the multi-drone surveillance problem. 
We first compare its performance in terms of ground-truth reward with that of a conventional expectimax approach in a simplified simulation. 
We then demonstrate the framework in two realistic simulated environments to examine the behaviour of MI-UCB in practical applications. 

\subsection{Comparison with Expectimax} 

We first compare the MI-UCB approach with the standard expectimax approach.
Here, expectimax refers to maximising the expected reward, given the current belief at each stage, without accounting for information gain. 

\begin{figure}[tb]
\centering
    \subfloat[Ground truth]{\includegraphics[width=0.49\columnwidth]{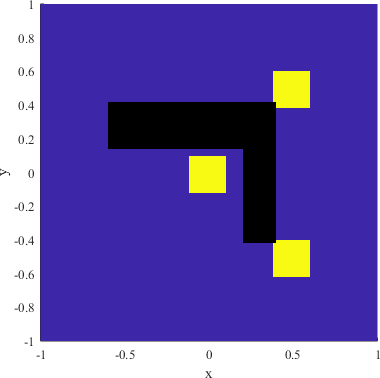}\label{fig:multi_drone_toy:ground_truth}}
    \subfloat[Percentage of targets confirmed.]{\includegraphics[width=0.49\columnwidth]{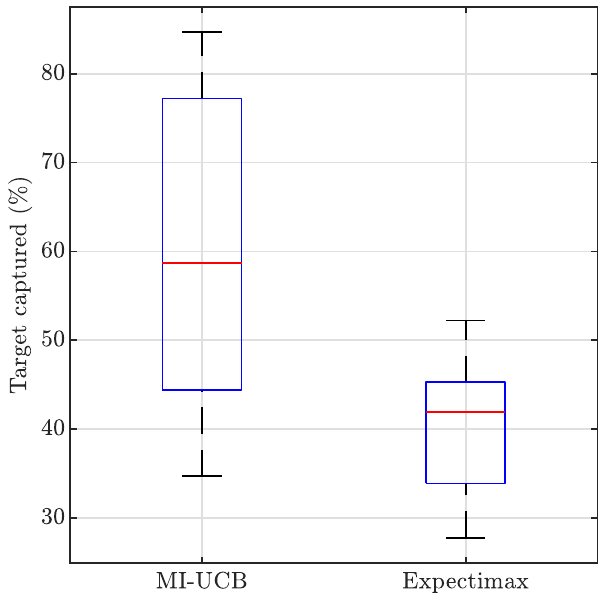}\label{fig:multi_drone_toy:comparison}} \\
    \subfloat[MI-UCB]{\includegraphics[width=0.49\columnwidth]{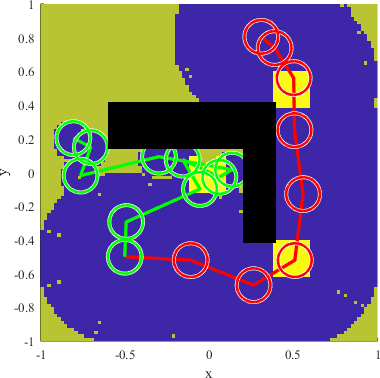}\label{fig:multi_drone_toy:mi_ucb}}
    \subfloat[Expectimax]{\includegraphics[width=0.49\columnwidth]{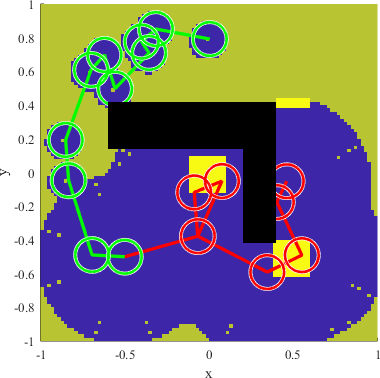}\label{fig:multi_drone_toy:expectimax}} 
\caption{Comparison between MI-UCB and expectimax in a simplified scenario. A robot confirms a target (yellow) if it is within the circle representing its target sensors' field of view (red for a scout-and-task drone equipped with target and long-range sensors and green for a task-only drone, equipped only with task sensors). The colourmap shows the belief on target occupancy (increasing from blue to yellow), while  black areas indicate obstacles and yellow-green areas are yet to be explored. MI-UCB (c) outperforms expectimax (d), because the former accounts for the fact that the red robot can provide greater information gain than the green robot.}
\vspace{-1ex}
\label{fig:multi_drone_toy}
\end{figure}

The comparison is set in the environment shown in Fig.~\ref{fig:multi_drone_toy:ground_truth}, where known obstacles and unknown targets are shown in black and yellow, respectively.
The task is to confirm targets within a given radius of a robot representing its task sensors' field of view.
A scout robot may also reveal knowledge about the environment using longer-range sensors.

There are two robots: red and green. 
To make the comparison fair, the red robot is a scout-and-task robot, while the green robot is task-only.
Thus, both are task robots, so the expectimax approach can generate a meaningful plan for each.
If one robot were to be scout-only, the expectimax approach would not generate a plan for it, unlike MI-UCB (Sec.~\ref{sec:mi-ucb}).
Intuitively, the expectimax approach simply reacts to the updates in belief, while MI-UCB accounts for information gain associated with the belief update.

The robots start with a uniform prior; and the robots' trajectory length for each time step is fixed at 2.5~m. 
Each robot updates its environment belief after executing one time step and re-plans its trajectory. 
We measure the performance in terms of the fraction of targets confirmed in simulation runs with the MI-UCB or expectimax approach generating the robots' trajectories, while randomising the environment by placing (three) targets in different locations for each run.

Combined results for ten runs of each simulation, provided in Fig.~\ref{fig:multi_drone_toy:comparison}, demonstrate that the MI-UCB approach outperforms the expectimax approach by $\sim50\%$ in terms of the median fraction of targets confirmed.
Examples that illustrate this trend are shown in Figs.~\ref{fig:multi_drone_toy:mi_ucb} and~\ref{fig:multi_drone_toy:expectimax}. 
In Fig.~\ref{fig:multi_drone_toy:mi_ucb}, it may be observed that MI-UCB causes the red scout-and-task robot to (in effect) `delegate' the task of confirming the target in the centre of the environment to the green task-only drone, unlike the expectimax approach used to generate the results shown in Fig.~\ref{fig:multi_drone_toy:expectimax}. 
This is because information gain is considered in MI-UCB. Therefore, the team can maximise its utility if the red robot continues to explore and gather information, while the green robot confirms the target. 
In contrast, in the expectimax approach, there is no incentive to do so; and the red scout-and-task robot, being closer, confirms the target in the centre. 
The higher variance of the MI-UCB approach is attributed to its optimism and shows that the upper limit of attainable reward is increased by valuing exploration.

\begin{figure}[t!]
    \centering
    \subfloat[Reward per unit distance\label{fig:composition:reward_per_distance}]{\includegraphics[width=0.49\columnwidth]{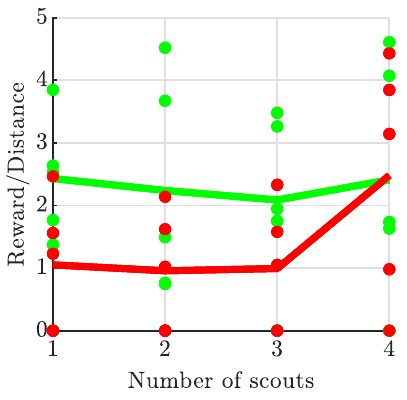}}
    \subfloat[Performance improvement\label{fig:composition:improvement}]{\includegraphics[width=0.49\columnwidth]{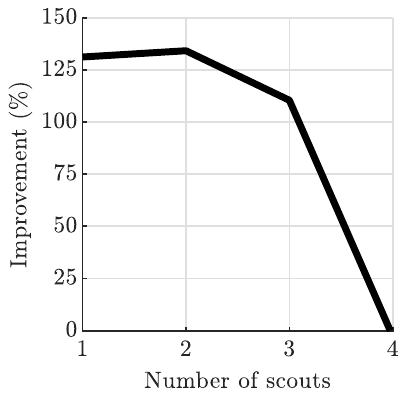}}
    \caption{(a) Reward obtained with MI-UCB (green) and expectimax (red) as a function of the number of scouts in the four-robot scenario. (b) MI-UCB provides the most benefit compared with expectimax with two scouts, an improvement of 134\%, and converges to equivalence with expectimax when four scouts are used.}
    \label{fig:composition}
    \vspace{-1ex}
\end{figure}
We verify this trend in comparative performance with a four-robot experiment. 
As before, all drones are task drones, but the number of scout-and-task drones is varied from one to four.
We fix the number of time steps and evaluate the reward per unit distance travelled by the drones in five runs for each approach, with five randomly placed targets.

\begin{figure*}[t!]
\centering
    \subfloat[]{\includegraphics[height=0.243\textwidth]{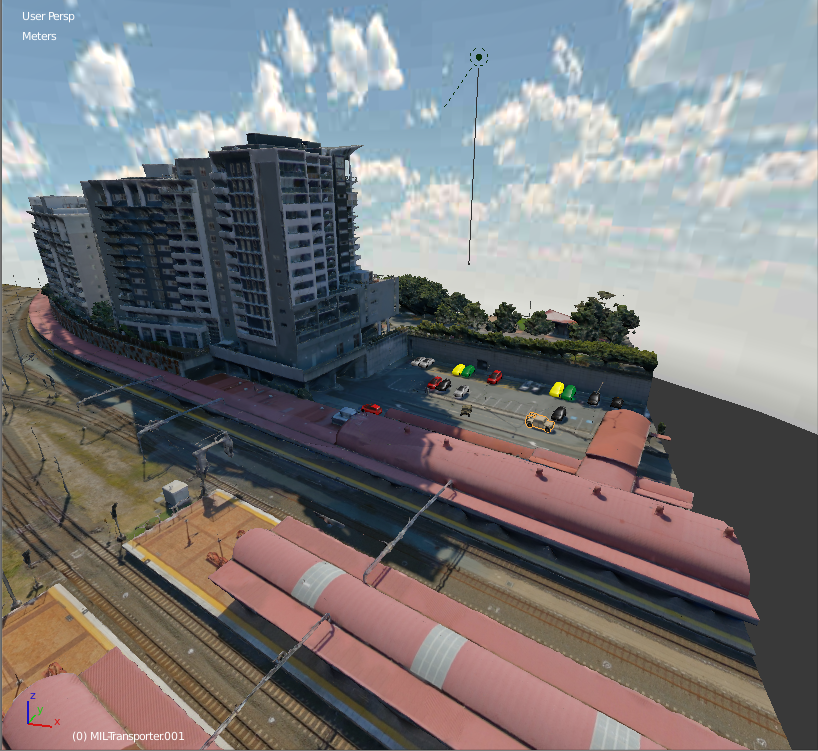}\label{fig:brisbane_env}}    
    \subfloat[]{\includegraphics[width=0.24\textwidth]{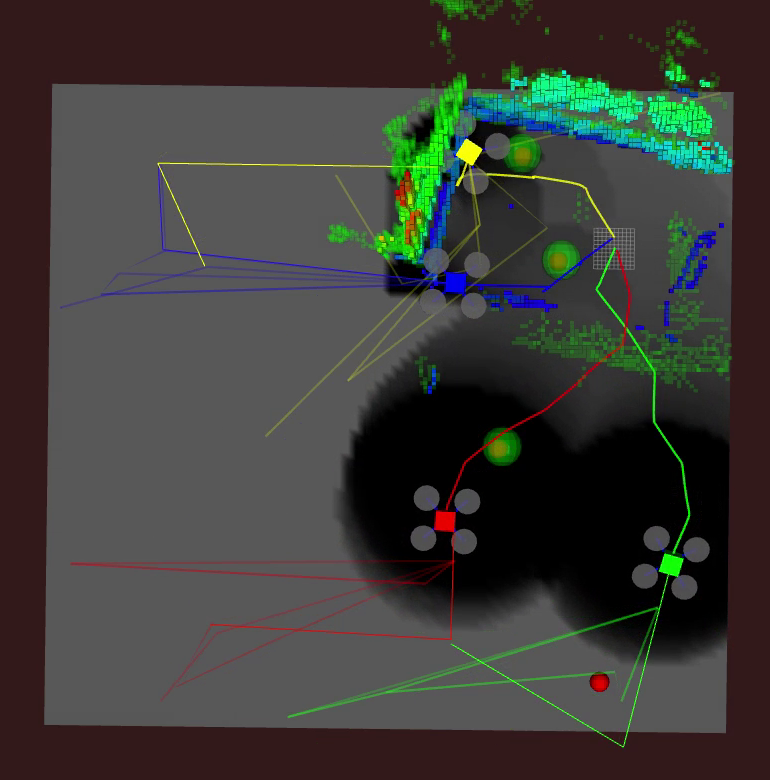}\label{fig:brisbane_2}} 
    \subfloat[]{\includegraphics[width=0.24\textwidth]{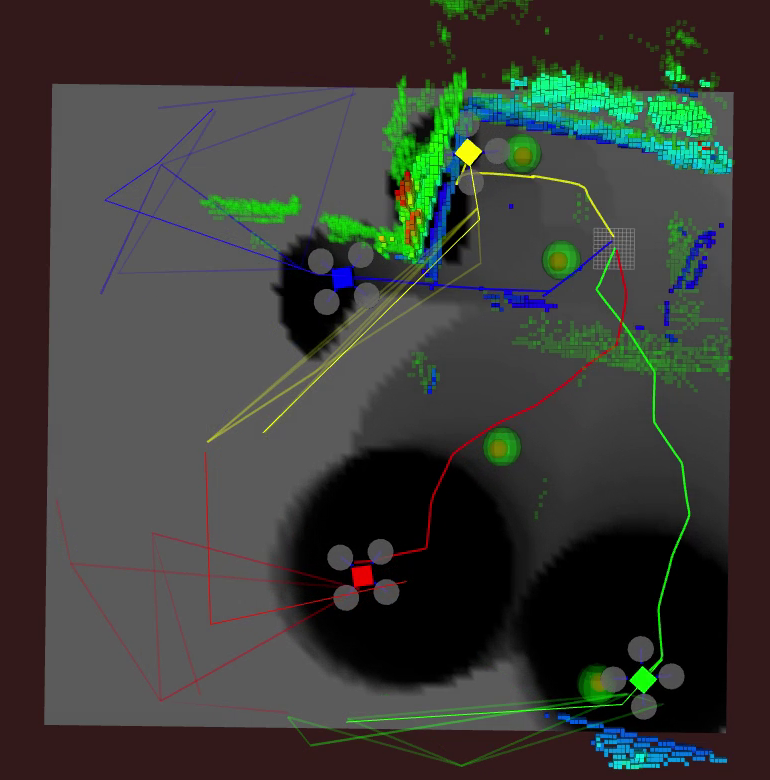}\label{fig:brisbane_3}} 
    \subfloat[]{\includegraphics[width=0.24\textwidth]{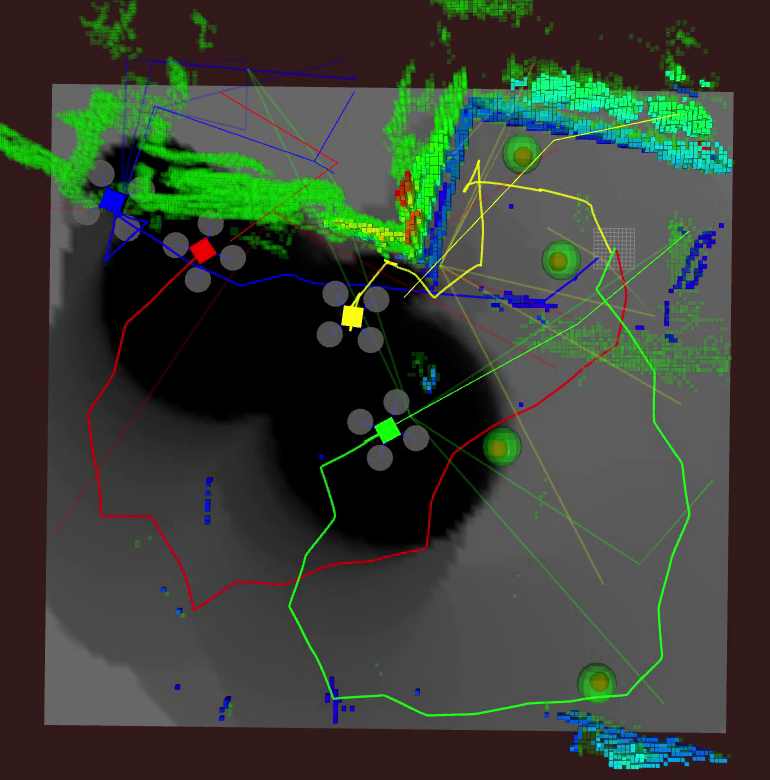}\label{fig:brisbane_4}} \\
    \subfloat[]{\includegraphics[height=0.243\textwidth]{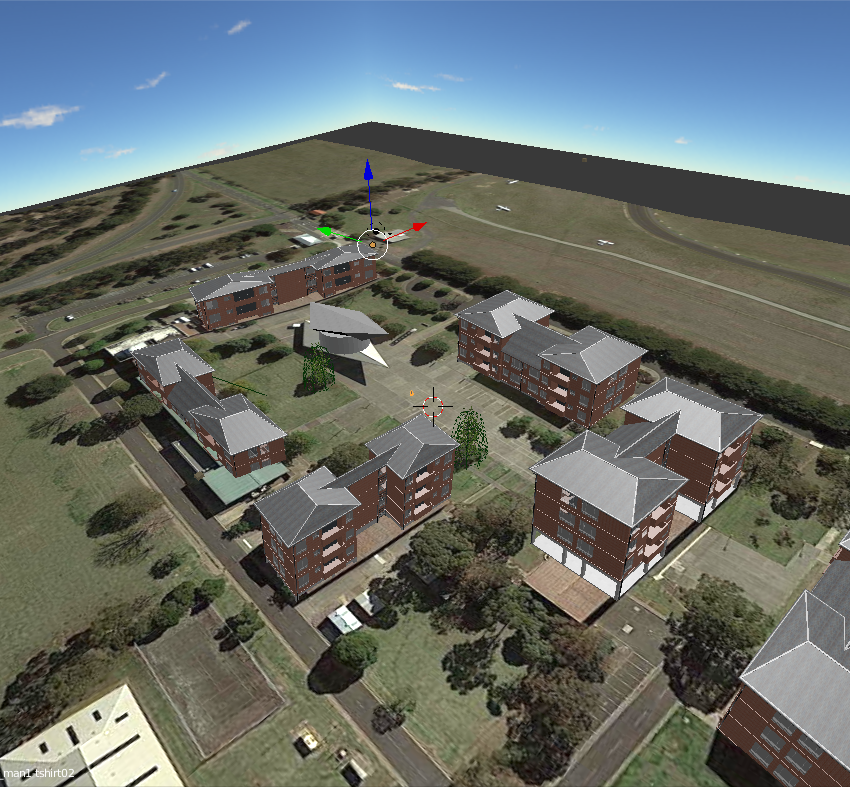}\label{fig:ptcook_env}} 
    \subfloat[]{\includegraphics[width=0.24\textwidth]{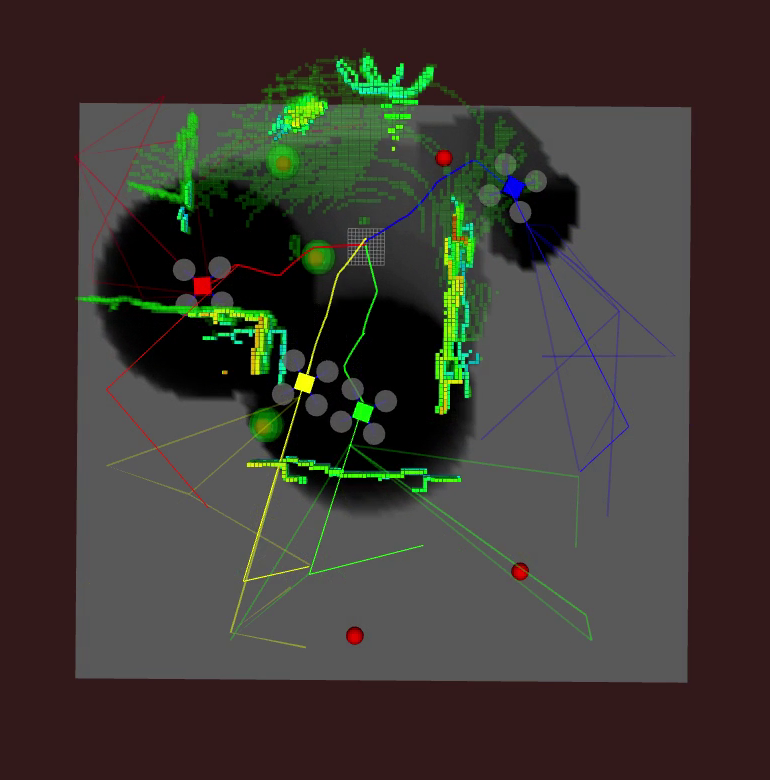}\label{fig:ptcook_2}} 
    \subfloat[]{\includegraphics[width=0.24\textwidth]{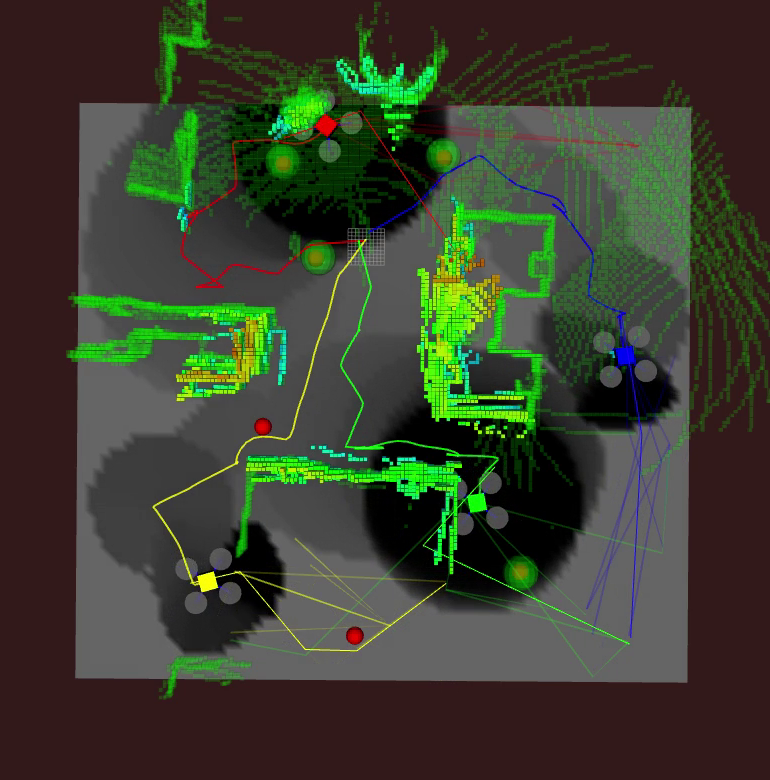}\label{fig:ptcook_3}} 
    \subfloat[]{\includegraphics[width=0.24\textwidth]{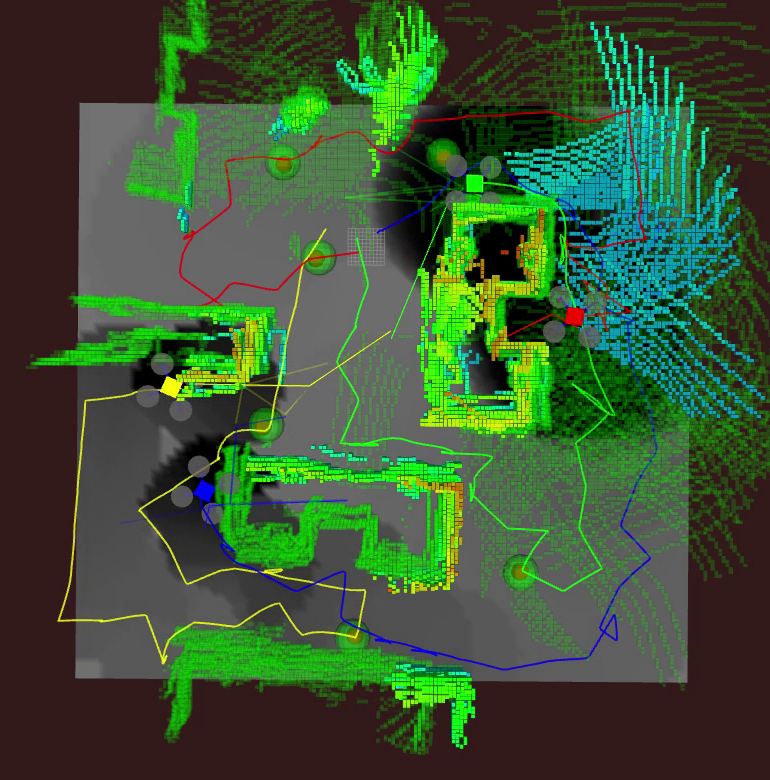}\label{fig:ptcook_4}}
\caption{Simulation results in the Brisbane (a) and Point Cook (e) environments. Time runs horizontally. Red spheres are targets yet to be confirmed. They turn green after confirmation. The grey colourmap shows the robot's belief over target occupancy.}
\label{fig:brisbane_rviz}
\vspace{-1ex}
\end{figure*}
As illustrated in Fig.~\ref{fig:composition}, when there are \emph{fewer} scout drones, MI-UCB provides greater performance benefit compared with expectimax; however, when all drones are scout-and-task drones, the approaches yield identical rewards.
This implies that MI-UCB makes better use of \emph{limited} information to provide consistent reward values with varying team composition. 
The performance improvement plateaus with two scouts, as shown in Fig.~\ref{fig:composition:improvement}, motivating the question of what team composition is \emph{optimal} for a given problem.

\subsection{Practical Demonstrations}
We demonstrate multi-drone surveillance in two realistic simulation environments. 
To examine practical efficacy, we perturb the problem from the ideal by varying the belief over time, introducing obstacles with simultaneous mapping, and emulating sensor failure.

We implement the multi-drone surveillance framework in performant software based on the Robot Operating System \cite{ros}.
Each drone builds its own map using Real-Time Appearance-Based Mapping \cite{labbe_rtabmap}; and the maps are combined to achieve inter-robot localisation and decentralised mapping.
We defer the description of the mapping framework to a future work. 
The grid-based filter for target estimation is implemented by use of the grid\_map library~\cite{grid_map_eth}.

The quadrotor simulation is based on the PX4 software-in-the-loop simulation \cite{meier_px4}, coupled with a modified version of the Modular Open Robotics Simulation Engine \cite{misssys_fork}.
The simulation is distributed over two desktop computers, each equipped with an NVIDIA RTX2060 graphics card. 
The computations for each drone are executed in real-time on an NVIDIA Jetson AGX single-board computer.

We first demonstrate the framework with four drones in the environment pictured in Fig.~\ref{fig:brisbane_env}. It is based on an urban area near Roma St. Station, Brisbane, Australia. 
The 3D model is generated by use of high-altitude photogrammetry and contains a mixture of open and cluttered terrain.

Figs.~\ref{fig:brisbane_2}\textendash \ref{fig:brisbane_4} show the target occupancy belief held by the green scout-and-task drone, as well as its intent. 
A target is identified at the start of the operation and is confirmed by the blue task drone (Fig.~\ref{fig:brisbane_rviz}b). 
The yellow task drone confirms another target in the upper-middle portion of the environment, where a parking lot is located.
This behaviour is consistent with the simplified simulation, in which, when cued by a scout drone, a task drone undertakes the task instead, distributing the exploration\textendash exploitation workload. 

Throughout the mission, the blue and yellow (task) drones focus on smaller, geometrically complex areas (around the tall building and the parking lot), while the red and green scout-and-task drones jointly cover a larger area above the train tracks. 
This demonstrates MI-UCB's inheritance from and generalisation of heterogeneous information gathering.

The difference between the two approaches is that MI-UCB results in all targets eventually being confirmed, thus completing the intended task.
A heterogeneous information-gathering approach would accept the long-range sensor's coverage of a target and not require that a short-range sensor capture it.
On the other hand, our decision-making under uncertainty approach allows the practitioner to specify that confirmation by a task drone is imperative.
This provides great flexibility, as one can easily replace the task of visual confirmation with, e.g., payload delivery or casualty evacuation, each of which requires proximity.

We also demonstrate the framework in the environment shown in Fig.~\ref{fig:ptcook_env}, which is modelled on RAAF Base Point Cook, Australia. 
The environment is prepared by modelling the buildings from a satellite image. 
It creates an interesting scenario for low-altitude operations because of the alleyways and corners that limit full visibility. 
In this simulation, we emulate perception failures to examine their effect on the performance of the algorithm. 
For example, in Fig.~\ref{fig:ptcook_2}, the blue scout-and-task drone has confirmed a target, but that is not reported to the other drones; 
while in Fig.~\ref{fig:ptcook_3}, the same occurs with yellow drone. 
Despite these unmodelled failures, the algorithm adapts to the change and successfully confirms all targets eventually, as illustrated in Fig.~\ref{fig:ptcook_4}.

\section{Conclusion}
We presented a framework for coordinating a scout\textendash task robot team undertaking exploration and exploitation simultaneously and synergistically. This behaviour is enabled by MI-UCB, a novel upper confidence bound that leads to increased task performance in hindsight compared to simply maximising the expected reward given the current belief. 
The generality and simplicity of MI-UCB motivates not only complex problems such as temporal-logic synthesis \cite{chanyeol_probabilistic,takashi_tanaka,brian_stl}, but also new fundamental questions in multi-robot coordination.
Given a problem instance, can we postulate an optimal composition of scout and task robots?
Can the composition be adapted dynamically depending on the task at hand? 
These types of coordination problems have many practical applications in areas such agriculture, infrastructure monitoring, construction, marine robotics, and others where there is value in collecting detailed observations of objects of interest that are distributed within the environment at unknown positions.

\bibliography{references}

\begin{thebibliography}{10}
\providecommand{\url}[1]{#1}
\csname url@rmstyle\endcsname
\providecommand{\newblock}{\relax}
\providecommand{\bibinfo}[2]{#2}
\providecommand\BIBentrySTDinterwordspacing{\spaceskip=0pt\relax}
\providecommand\BIBentryALTinterwordstretchfactor{4}
\providecommand\BIBentryALTinterwordspacing{\spaceskip=\fontdimen2\font plus
\BIBentryALTinterwordstretchfactor\fontdimen3\font minus
  \fontdimen4\font\relax}
\providecommand\BIBforeignlanguage[2]{{%
\expandafter\ifx\csname l@#1\endcsname\relax
\typeout{** WARNING: IEEEtran.bst: No hyphenation pattern has been}%
\typeout{** loaded for the language `#1'. Using the pattern for}%
\typeout{** the default language instead.}%
\else
\language=\csname l@#1\endcsname
\fi
#2}}

\bibitem{amazon}
P.~R. Wurman, R.~D'Andrea, and M.~Mountz, ``Coordinating hundreds of
  cooperative, autonomous vehicles in warehouses,'' \emph{AI Mag.}, vol.~29,
  no.~1, pp. 9--9, 2008.

\bibitem{swarmfarm}
G.~D’Urso, S.~L. Smith, R.~Mettu, T.~Oksanen, and R.~Fitch, ``Multi-vehicle
  refill scheduling with queueing,'' \emph{Comput. and Electron. in
  Agriculture}, vol. 144, pp. 44 -- 57, 2018.

\bibitem{a2ad-with-swarming}
R.~Gorrell, ``Countering {A2/AD} with swarming,'' Ph.D. dissertation, AIR
  UNIVERSITY, 2016.

\bibitem{ali_agha_mohamadi_where_to_map}
T.~Sasaki, K.~Otsu, R.~Thakker, S.~Haesaert, and A.~Agha-mohammadi, ``Where to
  map? {I}terative rover-copter path planning for {M}ars exploration,''
  \emph{Robot. and Automat. Lett.}, vol.~5, no.~2, pp. 2123--2130, 2020.

\bibitem{mohammadi_2}
K.~Ebadi and A.~Agha-Mohammadi, ``Rover localization in {M}ars helicopter
  aerial maps: Experimental results in a {M}ars-analogue environment,'' in
  \emph{Proc. of Int. Symp. on Exp. Robot. (ISER)}, 2018.

\bibitem{takashi_tanaka}
S.~{Bharadwaj}, M.~{Ahmadi}, T.~{Tanaka}, and U.~{Topcu}, ``Transfer entropy in
  {MDP}s with temporal logic specifications,'' in \emph{Proc. of Conf. on
  Decis. and Contr. (CDC)}, 2018, pp. 4173--4180.

\bibitem{decmcts}
G.~Best, O.~M. Cliff, T.~Patten, R.~R. Mettu, and R.~Fitch, ``Dec-{MCTS}:
  Decentralized planning for multi-robot active perception,'' \emph{The Int. J.
  of Robot. Res.}, vol.~38, no. 2-3, pp. 316--337, 2019.

\bibitem{markets}
M.~B. Dias, R.~Zlot, N.~Kalra, and A.~Stenz, ``Market-based multirobot
  coordination: A survey and analysis,'' \emph{Proc. of the IEEE}, vol.~94,
  no.~7, pp. 1257--1270, 2006.

\bibitem{hungarian}
G.~A. Korsah, A.~Stentz, and M.~B. Dias, ``A comprehensive taxonomy for
  multi-robot task allocation,'' \emph{The Int. J. of Robot. Res.}, vol.~32,
  no.~12, pp. 1495--1512, 2013.

\bibitem{smith_and_hollinger}
A.~J. Smith, G.~Best, J.~Yu, and G.~A. Hollinger, ``Real-time distributed
  non-myopic task selection for heterogeneous robotic teams,'' \emph{Auton.
  Robots}, vol.~43, no.~3, pp. 789--811, 2019.

\bibitem{hollinger_survey}
T.~H. Chung, G.~A. Hollinger, and V.~Isler, ``Search and pursuit-evasion in
  mobile robotics,'' \emph{Auton. Robots}, vol.~31, no.~4, p. 299, 2011.

\bibitem{vidal_pursuit_evasion}
R.~{Vidal}, O.~{Shakernia}, H.~J. {Kim}, D.~H. {Shim}, and S.~{Sastry},
  ``Probabilistic pursuit-evasion games: Theory, implementation, and
  experimental evaluation,'' \emph{Trans. on Robot. and Automat.}, vol.~18,
  no.~5, pp. 662--669, 2002.

\bibitem{clark2006gm}
D.~E. Clark, K.~Panta, and B.-N. Vo, ``The {GM-PHD} filter multiple target
  tracker,'' in \emph{Int. Conf. on Info. Fusion}.\hskip 1em plus 0.5em minus
  0.4em\relax IEEE, 2006, pp. 1--8.

\bibitem{dames}
P.~Dames, P.~Tokekar, and V.~Kumar, ``Detecting, localizing, and tracking an
  unknown number of moving targets using a team of mobile robots,'' \emph{The
  Int. J. of Robot. Res.}, vol.~36, no. 13-14, pp. 1540--1553, 2017.

\bibitem{Sung2018}
Y.~{Sung} and P.~{Tokekar}, ``Algorithm for searching and tracking an unknown
  and varying number of mobile targets using a limited {FoV} sensor,'' in
  \emph{Proc. of IEEE ICRA}, 2017, pp. 6246--6252.

\bibitem{pursuit_evasion_uav_ugv}
R.~{Vidal}, S.~{Rashid}, C.~{Sharp}, O.~{Shakernia}, {Jin Kim}, and
  S.~{Sastry}, ``Pursuit-evasion games with unmanned ground and aerial
  vehicles,'' in \emph{Proc. of IEEE ICRA}, vol.~3, 2001, pp. 2948--2955 vol.3.

\bibitem{optimal_search_for}
F.~Bourgault, T.~Furukawa, and H.~Durrant-Whyte, ``Optimal search for a lost
  target in a {B}ayesian world,'' \emph{Springer Tracts in Adv. Robot.},
  vol.~24, pp. 209--222, 01 2003.

\bibitem{hollinger_multi_robot_moving}
G.~Hollinger, S.~Singh, J.~Djugash, and A.~Kehagias, ``Efficient multi-robot
  search for a moving target,'' \emph{The Int. J. of Robot. Res.}, vol.~28,
  no.~2, pp. 201--219, 2009.

\bibitem{seng_keat_gan_collision}
S.~K. {Gan}, R.~{Fitch}, and S.~{Sukkarieh}, ``Real-time decentralized search
  with inter-agent collision avoidance,'' in \emph{Proc. of IEEE ICRA}, 2012,
  pp. 504--510.

\bibitem{brent_schotfeldt}
Y.~Kantaros, B.~Schlotfeldt, N.~Atanasov, and G.~J. Pappas, ``Asymptotically
  optimal planning for non-myopic multi-robot information gathering,'' in
  \emph{Proc. of RSS}, FreiburgimBreisgau, Germany, June 2019.

\bibitem{cliff2018robotic}
O.~M. Cliff, D.~L. Saunders, and R.~Fitch, ``Robotic ecology: Tracking small
  dynamic animals with an autonomous aerial vehicle,'' \emph{Sci. Robot.},
  vol.~3, 2018.

\bibitem{auer}
P.~Auer, N.~Cesa-Bianchi, and P.~Fischer, ``Finite-time analysis of the
  multiarmed bandit problem,'' \emph{Mach. Learn.}, vol.~47, pp. 235--256, 05
  2002.

\bibitem{kl_ucb}
A.~Garivier and O.~Cappé, ``The {KL-UCB} algorithm for bounded stochastic
  bandits and beyond,'' in \emph{Proc. of Conf. Comput. Learn. Theory}, 2011,
  pp. 359--376.

\bibitem{davidsilver}
D.~Silver and J.~Veness, ``{Monte-Carlo} planning in large {POMDPs},'' in
  \emph{Adv. in Neural Info. Process. Syst.}, J.~Lafferty, C.~Williams,
  J.~Shawe-Taylor, R.~Zemel, and A.~Culotta, Eds., vol.~23.\hskip 1em plus
  0.5em minus 0.4em\relax Curran Associates, Inc., 2010.

\bibitem{gp_ucb}
N.~Srinivas, A.~Krause, S.~Kakade, and M.~Seeger, ``Gaussian process
  optimization in the bandit setting: No regret and experimental design,'' in
  \emph{Proc. of Int. Conf. on Mach. Learn.}, 07 2010, pp. 1015--1022.

\bibitem{brian_acra}
K.~M.~B. Lee, J.~J.~H. Lee, C.~Yoo, B.~Hollings, and R.~Fitch, ``Active
  perception for plume source localisation with underwater gliders,'' in
  \emph{Australas. Conf. on Robot. and Automat. (ACRA)}, 2018.

\bibitem{sungjoon_choi}
H.~{Ahn}, Y.~{Oh}, S.~{Choi}, C.~J. {Tomlin}, and S.~{Oh}, ``Online learning to
  approach a person with no regret,'' \emph{Robot. and Automat. Lett.}, vol.~3,
  no.~1, pp. 52--59, 2018.

\bibitem{stanford_guys}
X.~Lu and B.~V. Roy, ``Information-theoretic confidence bounds for
  reinforcement learning,'' in \emph{Adv. Neural Info. Process. Syst.},
  H.~Wallach, H.~Larochelle, A.~Beygelzimer, F.~d'Alch\'{e} Buc, E.~Fox, and
  R.~Garnett, Eds., vol.~32.\hskip 1em plus 0.5em minus 0.4em\relax Curran
  Associates, Inc., 2019.

\bibitem{despot}
A.~Somani, N.~Ye, D.~Hsu, and W.~S. Lee, ``{DESPOT}: Online {POMDP} planning
  with regularization,'' in \emph{Adv. in Neural Info. Process. Syst.},
  C.~J.~C. Burges, L.~Bottou, M.~Welling, Z.~Ghahramani, and K.~Q. Weinberger,
  Eds., vol.~26.\hskip 1em plus 0.5em minus 0.4em\relax Curran Associates,
  Inc., 2013.

\bibitem{donsker_and_varadhan}
M.~D. Donsker and S.~S. Varadhan, ``Asymptotic evaluation of certain {M}arkov
  process expectations for large time,'' \emph{Commun. on Pure and Appl.
  Math.}, vol.~36, p. 183–212, 1983.

\bibitem{dupuis}
P.~Dupuis and R.~S. Ellis, \emph{A Weak Convergence Approach to the Theory of
  Large Deviations}.\hskip 1em plus 0.5em minus 0.4em\relax Wiley, 1997.

\bibitem{seldin}
Y.~{Seldin}, F.~{Laviolette}, N.~{Cesa-Bianchi}, J.~{Shawe-Taylor}, and
  P.~{Auer}, ``{PAC-Bayesian} inequalities for martingales,'' \emph{IEEE Trans.
  on Info. Theory}, vol.~58, no.~12, pp. 7086--7093, 2012.

\bibitem{brafman}
R.~I. Brafman and M.~Tennenholtz, ``{R-MAX}: {A} general polynomial time
  algorithm for near-optimal reinforcement learning,'' \emph{J. of Mach. Learn.
  Res.}, vol.~3, no. Oct, pp. 213--231, 2002.

\bibitem{sebastian_thrun}
S.~Thrun, ``Learning occupancy grids with forward models,'' in \emph{Proc. of
  IROS}, vol.~3.\hskip 1em plus 0.5em minus 0.4em\relax IEEE, 2001, pp.
  1676--1681.

\bibitem{ros}
\BIBentryALTinterwordspacing
{Stanford Artificial Intelligence Laboratory et al.}, ``Robotic operating
  system.'' [Online]. Available: \url{https://www.ros.org}
\BIBentrySTDinterwordspacing

\bibitem{labbe_rtabmap}
M.~Labbé and F.~Michaud, ``{RTAB-Map} as an open-source lidar and visual
  simultaneous localization and mapping library for large-scale and long-term
  online operation,'' \emph{J. of Field Robot.}, vol.~36, 10 2018.

\bibitem{grid_map_eth}
P.~Fankhauser and M.~Hutter, ``A universal grid map library: Implementation and
  use case for rough terrain navigation,'' in \emph{Robot Operating System
  (ROS) – The Complete Reference}, A.~Koubaa, Ed.\hskip 1em plus 0.5em minus
  0.4em\relax Springer, 2016, vol.~1, ch.~5.

\bibitem{meier_px4}
F.~Furrer, M.~Burri, M.~Achtelik, and R.~Siegwart, ``{RotorS}---a modular
  {G}azebo {MAV} simulator framework,'' in \emph{Robot Operating System
  ({ROS}): The Complete Reference}, A.~Koubaa, Ed.\hskip 1em plus 0.5em minus
  0.4em\relax Springer, 2016, vol.~1, pp. 595--625.

\bibitem{misssys_fork}
\BIBentryALTinterwordspacing
{Mission Systems Pty. Ltd}, ``{MORSE}: the modular open robots simulator
  engine.'' [Online]. Available:
  \url{https://github.com/mission-systems-pty-ltd/morse}
\BIBentrySTDinterwordspacing

\bibitem{chanyeol_probabilistic}
C.~Yoo and C.~Belta, ``Control with probabilistic signal temporal logic,''
  \emph{arXiv preprint arXiv:1510.08474}, 2015.

\bibitem{brian_stl}
K.~M.~B. Lee, C.~Yoo, and R.~Fitch, ``Signal temporal logic synthesis as
  probabilistic inference,'' in \emph{Proc. of IEEE ICRA}, 2021.

\end{thebibliography}
\flushend
\end{document}